\documentclass[10pt, a4paper]{article}
\usepackage{lrec}
\usepackage{multibib}
\newcites{languageresource}{Language Resources}
\usepackage{graphicx}
\usepackage{tabularx}
\usepackage{soul}
\usepackage{subfig}

\usepackage{epstopdf}
\usepackage[utf8]{inputenc}

\usepackage{hyperref}
\usepackage{xstring}

\usepackage{color}

\title{Understanding Spatial Relations through Multiple Modalities }

\name{Soham Dan, Hangfeng He, Dan Roth}

\address{University of Pennsylvania \\
 \{sohamdan, hangfeng, danroth\}@seas.upenn.edu\\}

\abstract{
Recognizing spatial relations and reasoning about them is essential in multiple applications including navigation, direction giving and human-computer interaction in general. Spatial relations between objects can either be \textit{explicit} -- expressed as spatial prepositions, or \textit{implicit} -- expressed by spatial verbs such as \textit{moving}, \textit{walking}, \textit{shifting}, etc. Both these, but implicit relations in particular, require significant common sense understanding. In this paper, we introduce the task of inferring \textit{implicit} and \textit{explicit} spatial relations between two entities in an image. We design a model that uses both textual and visual information to predict the spatial relations, making use of both positional and size information of objects and image embeddings. We contrast our spatial model with powerful language models and show how our modeling complements the power of these, improving prediction accuracy and coverage and facilitates dealing with unseen subjects, objects and relations.   \\ \newline  }

\begin{document}

\maketitleabstract

\section{Introduction}
Humans are able to do common sense reasoning across a variety of modalities -- textual, visual and for a variety of tasks -- reasoning, locating, navigation.
Several such tasks require spatial knowledge understanding and reasoning \cite{kordjamshidi2010spatial}, \cite{kordjamshidi2011spatial}, \cite{johnson2017clevr}, \cite{wang2017naturalizing}, \cite{baldridge2018points}. Although, there has been several recent works on common-sense reasoning ~\cite{speer2012representing}, \cite{emami2018knowledge}, \cite{sap2018atomic}, \cite{storks2019commonsense}, progress on spatial understanding and common-sense is rather limited. Prior work has either not been spatially focused ~\cite{lin2016leveraging}, \cite{xu2015show},\cite{antol2015vqa} or very restrictive in the class of spatial relations they handle ~\cite{xu2017commonsense}, \cite{kordjamshidi2017spatial}.
Recently, \cite{collell2018acquiring} presented the task of predicting an object's location and size in an image given the subject's bounding box and the spatial relation between them. 
\begin{figure}
  \includegraphics[width=\linewidth]{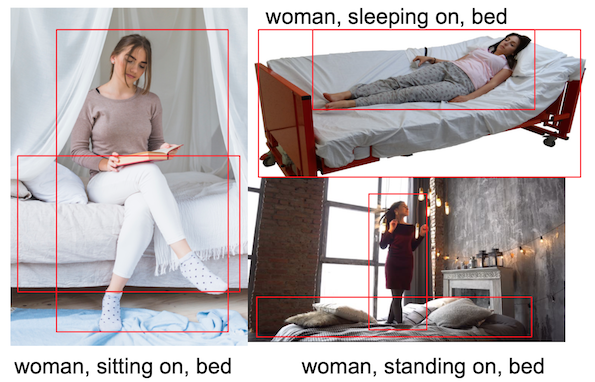}
  \caption{woman [?] bed. Language models adopt the choice seen most commonly, i.e., sleeping on, but we propose an image-specific model.}
  \label{fig:bed}
\end{figure}
\\
In this paper, we address the problem of understanding spatial relations. We specifically want to infer the spatial relationship between two entities given an image involving them. Spatial relations can either be -- \textit{explicit} (spatial prepositions such as \textit{on}, \textit{above}, \textit{under}) or  \textit{implicit} (intrinsic spatial concepts associated with actions -- \textit{sleeping}, \textit{sitting}, \textit{flying}). We take as input an image and the bounding boxes of the entities which are spatially related, the word and image embeddings of the two entities, and we want to predict the spatial relation between them (eg: \textit{sleeping}, \textit{standing}, \textit{sitting-on} in Figure \ref{fig:bed}. We compare this with powerful language models like BERT \cite{devlin2018bert} which have been trained on very large text corpora in a variety of contexts. Although BERT is not conditioned on the image, it can still provide a good list of candidate relations between the two entities using only the structured text information and it can provide new relations unseen in the specific dataset  we train on. We show that these complementary attributes --  being able to use the image-specific information by a task-specific model and  being able to predict a wide range of relations -- by BERT, can together give better performance than either approach alone, especially in low-resource and generalized settings ~\cite{collell2018acquiring}. For the \textit{woman,?,bed} example, if we were to rely on a language model alone, the prediction would be \textit{laying on} or \textit{sleeping on} almost all the time. This is because using just the textual modality makes it blind to the image specific cues. Using both the textual and visual information leads to a more robust model conditioned on both the image and the subject and object text description. 
\\
\textbf{Our contributions} are four-fold: \textbf{(1)} New task definition of \textit{explicit} and \textit{implicit} spatial relation prediction for two entities in an image. We explore another dimension of commonsense understanding of spatial relations with visual and language information. \textbf{(2)} Usage of \textbf{Spatial BERT}: combination of BERT and a spatial model. We conduct thorough experiments to show the role of the image, position and language information for the task under different settings-varying the number of training examples, the type of the spatial relations and the type of BERT.\footnote{https://github.com/sdan2/Multimodal-Spatial} \textbf{(3)} As a byproduct, we propose a re-scoring technique for evaluating this model combination. \textbf{(4)} We show that \textbf{Spatial BERT} is able to  predict in the generalized setting -- for unseen subjects, objects or relations.  

\section{Model Details}
\subsection{The Basics}
This task is to predict the spatial relation $R$ given the subject $S$ and the object $O$. For the subject $S$, we have the corresponding text information $T_s$, position information $P_s$, and image information $I_s$. Similarly, we have the corresponding text information $T_o$, position information $P_o$, and image information $I_o$ for the object $O$.
\subsection{The Spatial Model}
{\bf Feed Forward Network (FF).}
For the text information, we use average (for multi-word subjects and objects) glove embeddings \cite{pennington2014glove} of the words in the text to represent it. For simplicity, we use $w_s$ and $w_o$ to represent the average glove embbeddings for the $T_s$ and $T_o$. As for the position $P$, it contains $4$ float values denoting the $x$ and $y$ coordinate of the subject(object) center and the half-width and half-height of the bounding box of the subject(object). We then pass $w_s$, $w_o$, $P_s$, $P_o$ through a feed forward network with $128$ neurons and take a softmax of the predictions of all possible relations:
$$h = [\sigma(W_s[w_s; P_s] + b_s); \sigma(W_o[w_o; P_o] + b_o)]$$
$$\hat{p} = softmax(W_h h + b_h)$$
where $\sigma$ is the activation function ReLU.
\\
{\bf Feed Forward Network + Image Embeddings (FF+I).} 
In addition to the text information and position information of the subject and object, we use pre-trained visual embeddings to represent the image information $I_s$ and $I_o$. For simplicity, we use $i_s$ and $i_o$ to represent the visual embeddings for the subject and the object. The visual embeddings of the words are provided by ~\cite{collell2018learning} from a VGG128 network \cite{chatfield2014return} pre-trained on Imagenet and fine-tuned on Visual Genome ~\cite{krishnavisualgenome}. The hidden representation in the FF + Image Embeddings are as following:
$$h = [\sigma(W_s[w_s; P_s; i_s] + b_s); \sigma(W_o[w_o; P_o; i_o] + b_o)]$$

\subsection{The Language Model}
{\bf BERT.}
We use BERT \cite{devlin2018bert} as language model to predict the most likely spatial relation by masking the relation and providing "subject [MASK] object" as input and running the beam search to obtain top $20$ predictions. 
\\
{\bf Fine-tuned BERT (f-BERT).}
We fine-tune BERT for the Visual Genome datatset by collecting all the "subject relation object" texts from the training data.
\subsection{Spatial BERT}
We combine the prediction of our best model (FF+I) with normal BERT and fine-tuned BERT to get two combined models. We essentially re-rank the predictions from the two models. Assume $\hat{p}_{bert}$ and $\hat{p}_{ff}$ are the predicted probability distribution from the BERT and FF+I models, the predicted probability of Spatial BERT is:
$\hat{p} = \hat{p}_{ff} + \lambda \hat{p}_{bert}$,
where $\lambda$ is a non-negative float to adjust the weight of the BERT predictions. 

\section{Experiments}
\subsection{Dataset}
\begin{figure}%
    \centering
    \subfloat[cat UNDER chair ]{{\includegraphics[height=0.2\textheight,width=.2\textwidth]{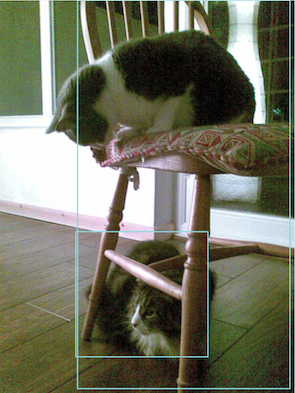} }}%
    \qquad
    \subfloat[cat ON chair]{{\includegraphics[height=0.2\textheight,width=.2\textwidth]{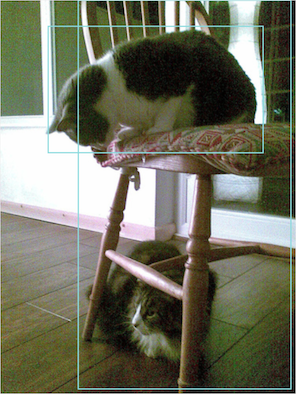} }}%
    \caption{Examples of explicit spatial relations from Visual Genome of $(cat, ?, chair)$}%
    \label{cats}%
\end{figure}

\begin{figure}%
    \centering
    \subfloat[man RIDING surfboard ]{{\includegraphics[height=0.2\textheight,width=.2\textwidth]{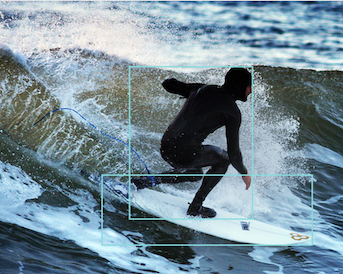} }}%
    \qquad
    \subfloat[man CARRYING surfboard]{{\includegraphics[height=0.2\textheight,width=.2\textwidth]{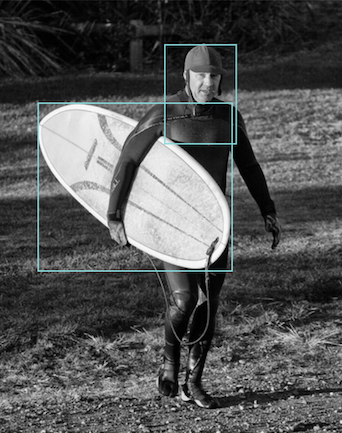} }}%
    \caption{Examples of implicit spatial relations from Visual Genome of $(man, ?, surfboard)$}%
    \label{surf}%
\end{figure}

We use the Visual Genome dataset ~\cite{krishnavisualgenome}. We work on the (subject, relation, object) triples where the subject and object is accompanied by the bounding box information\footnote{Note each image is scaled to $1$ so that the sizes are comparable. See ~\cite{collell2018acquiring} for more information on the data-preprocessing step.}. The dataset is partitioned into two categories: \textit{explicit} and \textit{implicit} based on the spatial relation in a triple and the experiments are performed separately for each category.  There are $333321$ implicit and $682374$ explicit triples. In Figure \ref{cats} we see examples from the dataset for the $explicit$ relations between $cat$ as $subject$ and $chair$ as $object$. In Figure \ref{surf} we see examples from the dataset for the $implicit$ relations between $man$ as $subject$ and $surfboard$ as $object$.

For the \textit{explicit} relations, \textit{on} is the majority relation with frequency $425308$ and {\bf the majority baseline for \textit{explicit} relation prediction is} ${\bf 62}${\bf\%}. Similarly, for the \textit{implicit} relations, \textit{has} is the majority relation with a frequency $167780$ and {\bf the majority baseline for \textit{implicit} relation prediction is} ${\bf 50}${\bf\%}. 
\subsection{GloVe based Re-Scoring Metric}
\begin{figure}
  \includegraphics[width=\linewidth]{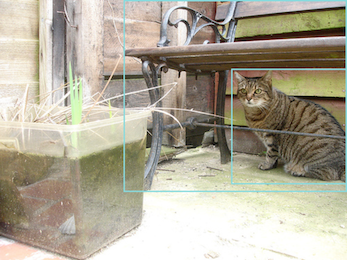}
  \caption{cat BENEATH chair. Even if we have never seen \textit{beneath} as a relation during the training phase but have seen cat UNDER chair (Figure \ref{cats}), using the BERT predictions and re-scoring the choices using the GloVe embedding similarity, we want to be able to predict unseen relations at test time.}
  \label{beneath}
\end{figure}

One of the principal benefits of using language models is that they can predict new relations never seen during the training of the spatial model (see Figure \ref{beneath}). In several cases (especially the low training data regimes)  where BERT is used in combination with the spatial model, BERT may predict spatial relations which are unseen in training and we need an effective strategy  to re-score the spatial model predictions based on these unseen predictions. For example, say, BERT predicts \textit{atop} (not seen during training) as the relation between \textit{book} and \textit{table} for an image whose gold triple is (\textit{book}, \textit{on}, \textit{table}). In such situations we develop a re-scoring metric to distribute the score BERT assigns to unseen relations (\textit{atop}) among the seen relations (\textit{on}, \textit{above}, \textit{over}) which are related to it. This relatedness is measured by the cosine similarity between the unseen word vector and the word vectors for relations present in the dataset. Thus, $score(on)=score(on)+ sim(on,atop)*score(atop)$.

\subsection{Experimental Settings}
We present the experiment results separately for the explicit and implicit relations. For each type of relation, we vary the percentage of data used for training as $1\%$, $10\%$, $50\%$, $75\%$, $100\%$. We try two variations of BERT -- the normal pre-trained BERT and f-BERT: BERT fine-tuned on the \textit{implicit} and \textit{explicit} dataset respectively. We try variety of $\lambda$ (from $0.01$ to $1$) to combine the BERT scores and the spatial model scores and we report the best result across different values of $\lambda$ ( we exclude $0$ or $\infty$ (large positive values) which are already reflected in only-BERT and only the spatial model results). In all the settings the train-development-test split is set as $75-15-15\%$.
\begin{table*}[t]
\begin{center}
\begin{tabular}{|c|c|c|c|c|c|c|c|c|c|c|} 
\hline
Spatial Relation  & 
              \multicolumn{5}{|c|}{Explicit} 
             
             & \multicolumn{5}{|c|}{Implicit}\\
\hline
$\%$ of Training Data &  1\% & 10\% & 50\% & 75\% & 100\%  & 1\% & 10\% & 50\% & 75\% & 100\% \\
 \hline
 BERT &  38.56 & 38.56 & 38.56 & 38.56 & 38.56 &  6.9  & 6.9 & 6.9 & 6.9 & 6.9\\ 
 f-BERT &  73.03 & 73.03 & 73.03 & 73.03 & 73.03 &  \textbf{77.6}  & \textbf{77.6} & 77.6 & 77.6 & 77.6\\  
 FF &  0.8 & 72.1 & 74 & 74.4 &74.7  &   0 & 75.0 & 78.8 & 79.0 & 79.5\\  
 FF+I &  0.71 & 72.7 & 74.1 & 74.5 & 74.72 &  0.01  & 75.79 & 78.86 & 79.3 & 79.5\\ 
 BERT+FF+I &  36.4 & 72.88 & \textbf{74.7} & \textbf{74.9} & \textbf{75.2} &  6.35 & 75.7 & 79 &79.3 & 79.5\\ 
 f-BERT+FF+I &  \textbf{73.06} & \textbf{73.06} & 74.2 & 74.52 & 74.74 & \textbf{77.6} & 77.5 & \textbf{79.3} & \textbf{79.7} & \textbf{79.9}\\ 
 
\hline

\end{tabular}
\end{center}
\caption{Comparison of performance (in percentage accuracy) of different models for explicit and implicit spatial relations, normal and f-BERT and combinations with the best spatial model for varying portions of training data.}
\end{table*}

\begin{table*}[t]
\centering
\begin{tabular}{|c|c|c|c|c|c|c|c| } 
\hline
Model & Expl(S, R) & Impl(S, R) & Expl(O, R) & Impl(O, R) & Expl(R) & Impl(R)\\
 \hline
 BERT & 61.9 & 14.4 &  59.9 & 27.0 & {\bf 24.1} & {\bf 13.7} \\ 
  
 FF+I & 60.1 & {\bf 68.6} &   59.5 & 50.1 & 0 & 0 \\ 
 BERT+FF+I & {\bf 67.4} & 59.8 &  {\bf 62.5} & {\bf 54.1} & 24.0 & {\bf 13.7}\\ 
 
 \hline
\end{tabular}
\caption{Experiments for unseen subjects, objects and relations (for both explicit and implicit relations). Spatial BERT gives better performance than BERT or FF+I for Impl(O, R) but not in Impl(S, R) potentially because the subject set is much sparser than the object set. }
\label{table:zero-shot}
\end{table*}

\subsection{Results}
We see that the best performance in several of the settings is achieved by Spatial BERT. Although the gains may look small, the number of data-points is huge and thus, the improvement is significant in terms of absolute counts. For smaller percentages of training data, the spatial models perform poorly but later on beats BERT. Also, notice BERT performs significantly better for the explicit spatial relations than the implicit spatial relations possibly because the set of implicit relations is much larger and the task of predicting them requires more image understanding and common-sense compared to the explicit spatial relations. BERT and f-BERT performances do not change across different data percentages because they are both just used for inference and the amount of training data does not affect them.

\section{Unseen Subject, Object or Relation }
It is greatly desirable that models learn to generalize to unseen contexts and is necessary for true spatial understanding of the relations. 
\subsection{The Settings}
\begin{figure}%
    \centering
    \subfloat[man RIDING elephant ]{{\includegraphics[height=0.2\textheight,width=.2\textwidth]{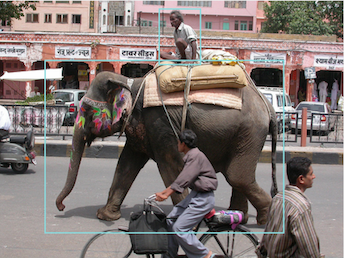} }}%
    \qquad
    \subfloat[man RIDING bike]{{\includegraphics[height=0.2\textheight,width=.2\textwidth]{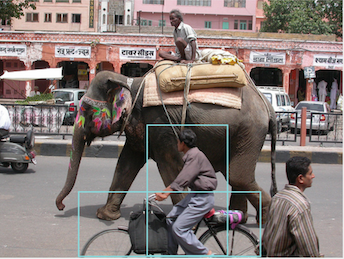} }}%
    \caption{In this example from Visual Genome, suppose we have seen $(man,riding,elephant)$ in training. In the second image, we want to be able to predict \textit{riding}, even if we have never seen the $(riding,bike)$ combination before}%
    \label{elebik}%
\end{figure}

If a spatial relation was seen during training -- (\textit{man}, \textit{riding}, \textit{horse}) and the supporting image, the model should be able to infer that the \textit{implicit} (in this example) spatial relation for a new image depicting (\textit{lady}, \textit{riding}, \textit{elephant}) should be \textit{riding}, even if it has never seen the subject (\textit{lady}) or object (\textit{elephant}) before. We show two illustrative examples from Visual Genome that we want to handle for the unseen subject (Figure \ref{manwoman}) and unseen object (Figure \ref{elebik}) settings respectively. The  pre-trained embeddings of the unseen subject(object) is similar to the embeddings of similar seen subject(object) and this (along with the positional information) should
help the model identify that the relations are similar.
To systematically test this capability, we perform experiments for the \textit{implicit} and \textit{explicit} relations in Table \ref{table:zero-shot}. For each dataset type we experiment with three settings -- \textbf{unseen subject, unseen object and unseen relation}. 

The unseen subject(object) setting is relatively easier compared to the unseen relation setting. For subject, we test if we can correctly predict the \textit{flying} relation in (\textit{kid}, \textit{flying}, \textit{kite}) even if we have never seen (\textit{kid}, \textit{flying}, $X$) during training. For objects, we test if we can correctly predict the \textit{riding} relation in (\textit{man}, \textit{riding}, \textit{elephant}) even if we have never seen ($X$, \textit{riding}, \textit{elephant}) during training. Here, $X$ denotes any object or subject, respectively.  We first tabulate all the (subject, relation), (object,relation) pairs  and we split this list into the test set pairs($15\%$) and the training and development set pairs($85\%$). We then form the test set (train, development) by collecting all data-points whose (subject,relation), (object,relation) is in the test( train or development) set pairs. Thus, the test set has (subject,relation), (object,relation) which are not seen during training. We only use the normal pre-trained BERT for the generalized experiments since fine-tuning is not very natural for these settings.

\begin{figure}%
    \centering
    \subfloat[man RIDING elephant ]{{\includegraphics[height=0.2\textheight,width=.2\textwidth]{manelep.png} }}%
    \qquad
    \subfloat[woman RIDING elephant]{{\includegraphics[height=0.2\textheight,width=.2\textwidth]{womanelep.png} }}%
    \caption{In this example from Visual Genome,suppose we have seen $(man,riding,elephant)$ in training. In the second image, we want to be able to predict \textit{riding}, even if we have never seen the $(woman,riding)$ combination before}%
    \label{manwoman}%
\end{figure}

\subsection{Analysis}
{\bf Unseen Subject.} For the explicit relations, we see that Spatial BERT performs much better than either model in isolation. This is potentially because the explicit relations are a smaller set and easier for BERT to predict and this in turn helps Spatial BERT. However, since the class of implicit relations are much larger and subtle BERT performs very poorly and spatial model by itself performs the best for the implicit relations.
\\
{\bf Unseen Object.} As shown in Table \ref{table:zero-shot}, for both the explicit and the implicit relations Spatial BERT gives the best performance although BERT by itself does not perform very well.
\\
{\bf Unseen Relation.}  This task is  possible because BERT is able to predict a much larger class of relations and using our GloVe-scoring metric we can decompose the scores of these relations across the known set of relations and this should bring the correct relations towards the top of the ranked list. We also relax the accuracy metric by counting a prediction as correct if the gold relation is in the top-$5$ predicted relations. In this setting, the spatial model by itself cannot predict any unseen relation and thus, BERT gives the best performance. 
\section{Discussion}
We presented the task of spatial relation prediction and developed models that use position information, visual embeddings and word embeddings to predict relations. Further we show that combining this model with BERT helps for low resource settings and generalization over unseen subjects, objects and relations. However, the visual embeddings for entities which are used in our models are of limited use in some situations. For example, it is hard for our models to distinguish between (\textit{man}, \textit{running}, \textit{dog}) and (\textit{man}, \textit{walking}, \textit{dog}) given (\textit{man}, --, \textit{dog}). In future we want to use more principled ways of incorporating image specific information to have an even more fine-grained spatial relation classification. We also want to develop an interactive system that does both the object position prediction task ~\cite{collell2018acquiring} and the spatial relation prediction task for a spatially-involved domain such as Blocks World \cite{bisk2016towards}.

\section{Acknowledgment}
This work was supported by Contract W911NF-15-1-0461 with the US Defense Advanced Research Projects Agency (DARPA) and the Army Research Office (ARO) and by Contract FA8750-19-2-0201 with the US Defense Advanced Research Projects Agency (DARPA). Approved for Public Release, Distribution Unlimited. The views expressed are those of the authors and do not reflect the official policy or position of the Department of Defense or the U.S. Government.

\section{References}

\bibliographystyle{lrec}
\bibliography{lrec2020W-xample-kc}

\begin{thebibliography}{}

\bibitem[\protect\citename{Antol \bgroup et al.\egroup }2015]{antol2015vqa}
Antol, S., Agrawal, A., Lu, J., Mitchell, M., Batra, D., Lawrence~Zitnick, C.,
  and Parikh, D.
\newblock (2015).
\newblock Vqa: Visual question answering.
\newblock In {\em Proceedings of the IEEE international conference on computer
  vision}, pages 2425--2433.

\bibitem[\protect\citename{Baldridge \bgroup et al.\egroup
  }2018]{baldridge2018points}
Baldridge, J., Bedrax-Weiss, T., Luong, D., Narayanan, S., Pang, B., Pereira,
  F., Soricut, R., Tseng, M., and Zhang, Y.
\newblock (2018).
\newblock Points, paths, and playscapes: Large-scale spatial language
  understanding tasks set in the real world.
\newblock In {\em Proceedings of the First International Workshop on Spatial
  Language Understanding}, pages 46--52.

\bibitem[\protect\citename{Bisk \bgroup et al.\egroup }2016]{bisk2016towards}
Bisk, Y., Marcu, D., and Wong, W.
\newblock (2016).
\newblock Towards a dataset for human computer communication via grounded
  language acquisition.
\newblock In {\em AAAI Workshop: Symbiotic Cognitive Systems}.

\bibitem[\protect\citename{Chatfield \bgroup et al.\egroup
  }2014]{chatfield2014return}
Chatfield, K., Simonyan, K., Vedaldi, A., and Zisserman, A.
\newblock (2014).
\newblock Return of the devil in the details: Delving deep into convolutional
  nets.
\newblock {\em arXiv preprint arXiv:1405.3531}.

\bibitem[\protect\citename{Collell and Moens}2018]{collell2018learning}
Collell, G. and Moens, M.-F.
\newblock (2018).
\newblock Learning representations specialized in spatial knowledge: Leveraging
  language and vision.
\newblock {\em Transactions of the Association of Computational Linguistics},
  6:133--144.

\bibitem[\protect\citename{Collell \bgroup et al.\egroup
  }2018]{collell2018acquiring}
Collell, G., Van~Gool, L., and Moens, M.-F.
\newblock (2018).
\newblock Acquiring common sense spatial knowledge through implicit spatial
  templates.
\newblock In {\em Thirty-Second AAAI Conference on Artificial Intelligence}.

\bibitem[\protect\citename{Devlin \bgroup et al.\egroup }2018]{devlin2018bert}
Devlin, J., Chang, M.-W., Lee, K., and Toutanova, K.
\newblock (2018).
\newblock Bert: Pre-training of deep bidirectional transformers for language
  understanding.
\newblock {\em arXiv preprint arXiv:1810.04805}.

\bibitem[\protect\citename{Emami \bgroup et al.\egroup
  }2018]{emami2018knowledge}
Emami, A., De~La~Cruz, N., Trischler, A., Suleman, K., and Cheung, J. C.~K.
\newblock (2018).
\newblock A knowledge hunting framework for common sense reasoning.
\newblock {\em arXiv preprint arXiv:1810.01375}.

\bibitem[\protect\citename{Johnson \bgroup et al.\egroup
  }2017]{johnson2017clevr}
Johnson, J., Hariharan, B., van~der Maaten, L., Fei-Fei, L., Lawrence~Zitnick,
  C., and Girshick, R.
\newblock (2017).
\newblock Clevr: A diagnostic dataset for compositional language and elementary
  visual reasoning.
\newblock In {\em Proceedings of the IEEE Conference on Computer Vision and
  Pattern Recognition}, pages 2901--2910.

\bibitem[\protect\citename{Kordjamshidi \bgroup et al.\egroup
  }2010]{kordjamshidi2010spatial}
Kordjamshidi, P., Van~Otterlo, M., and Moens, M.-F.
\newblock (2010).
\newblock Spatial role labeling: Task definition and annotation scheme.
\newblock In {\em LREC}.

\bibitem[\protect\citename{Kordjamshidi \bgroup et al.\egroup
  }2011]{kordjamshidi2011spatial}
Kordjamshidi, P., Van~Otterlo, M., and Moens, M.-F.
\newblock (2011).
\newblock Spatial role labeling: Towards extraction of spatial relations from
  natural language.
\newblock {\em ACM Transactions on Speech and Language Processing (TSLP)},
  8(3):4.

\bibitem[\protect\citename{Kordjamshidi \bgroup et al.\egroup
  }2017]{kordjamshidi2017spatial}
Kordjamshidi, P., Rahgooy, T., and Manzoor, U.
\newblock (2017).
\newblock Spatial language understanding with multimodal graphs using
  declarative learning based programming.
\newblock In {\em Proceedings of the 2nd Workshop on Structured Prediction for
  Natural Language Processing}, pages 33--43.

\bibitem[\protect\citename{Krishna \bgroup et al.\egroup
  }2016]{krishnavisualgenome}
Krishna, R., Zhu, Y., Groth, O., Johnson, J., Hata, K., Kravitz, J., Chen, S.,
  Kalantidis, Y., Li, L.-J., Shamma, D.~A., Bernstein, M., and Fei-Fei, L.
\newblock (2016).
\newblock Visual genome: Connecting language and vision using crowdsourced
  dense image annotations.

\bibitem[\protect\citename{Lin and Parikh}2016]{lin2016leveraging}
Lin, X. and Parikh, D.
\newblock (2016).
\newblock Leveraging visual question answering for image-caption ranking.
\newblock In {\em European Conference on Computer Vision}, pages 261--277.
  Springer.

\bibitem[\protect\citename{Pennington \bgroup et al.\egroup
  }2014]{pennington2014glove}
Pennington, J., Socher, R., and Manning, C.
\newblock (2014).
\newblock Glove: Global vectors for word representation.
\newblock In {\em Proceedings of the 2014 conference on empirical methods in
  natural language processing (EMNLP)}, pages 1532--1543.

\bibitem[\protect\citename{Sap \bgroup et al.\egroup }2018]{sap2018atomic}
Sap, M., LeBras, R., Allaway, E., Bhagavatula, C., Lourie, N., Rashkin, H.,
  Roof, B., Smith, N.~A., and Choi, Y.
\newblock (2018).
\newblock Atomic: An atlas of machine commonsense for if-then reasoning.
\newblock {\em arXiv preprint arXiv:1811.00146}.

\bibitem[\protect\citename{Speer and Havasi}2012]{speer2012representing}
Speer, R. and Havasi, C.
\newblock (2012).
\newblock Representing general relational knowledge in conceptnet 5.
\newblock In {\em LREC}, pages 3679--3686.

\bibitem[\protect\citename{Storks \bgroup et al.\egroup
  }2019]{storks2019commonsense}
Storks, S., Gao, Q., and Chai, J.~Y.
\newblock (2019).
\newblock Commonsense reasoning for natural language understanding: A survey of
  benchmarks, resources, and approaches.
\newblock {\em arXiv preprint arXiv:1904.01172}.

\bibitem[\protect\citename{Wang \bgroup et al.\egroup
  }2017]{wang2017naturalizing}
Wang, S.~I., Ginn, S., Liang, P., and Manning, C.~D.
\newblock (2017).
\newblock Naturalizing a programming language via interactive learning.
\newblock {\em arXiv preprint arXiv:1704.06956}.

\bibitem[\protect\citename{Xu \bgroup et al.\egroup }2015]{xu2015show}
Xu, K., Ba, J., Kiros, R., Cho, K., Courville, A., Salakhutdinov, R., Zemel,
  R., and Bengio, Y.
\newblock (2015).
\newblock Show, attend and tell: Neural image caption generation with visual
  attention.
\newblock {\em arXiv preprint arXiv:1502.03044}.

\bibitem[\protect\citename{Xu \bgroup et al.\egroup }2017]{xu2017commonsense}
Xu, F.~F., Lin, B.~Y., and Zhu, K.~Q.
\newblock (2017).
\newblock Commonsense locatednear relation extraction.
\newblock {\em arXiv preprint arXiv:1711.04204}.

\end{thebibliography}


\end{document}